\documentclass[letterpaper, 10 pt, conference]{ieeeconf}  %

\IEEEoverridecommandlockouts                              %

\usepackage{tikz}
\def\*#1{\mathbf{#1}}

\usepackage{amsmath}
\usepackage{amsfonts}
\usepackage{dsfont}
\usepackage{booktabs} %
\usepackage{multirow}
\usepackage{graphicx}
\usepackage{cleveref}

\usepackage{amssymb}
\usepackage{pifont}
\newcommand{\cmark}{\ding{51}}%
\newcommand{\xmark}{\ding{55}}%

\usepackage{siunitx}

\usepackage{lipsum}  
\usepackage{svg}

\definecolor{car}{HTML}{6496F5}
\definecolor{bicycle}{HTML}{64E6F5}
\definecolor{motorcycle}{HTML}{1E3C96}
\definecolor{truck}{HTML}{501EB4}
\definecolor{other-vehicle}{HTML}{0000FF}
\definecolor{person}{HTML}{FF1E1E}
\definecolor{bicyclist}{HTML}{FF28C8}
\definecolor{motorcyclist}{HTML}{961E5A}
\definecolor{road}{HTML}{FF00FF}
\definecolor{parking}{HTML}{FF96FF}
\definecolor{sidewalk}{HTML}{4B004B}
\definecolor{other-ground}{HTML}{AF004B}
\definecolor{building}{HTML}{FFC800}
\definecolor{fence}{HTML}{FF7832}
\definecolor{vegetation}{HTML}{00AF00}
\definecolor{trunk}{HTML}{873C00}
\definecolor{terrain}{HTML}{96F050}
\definecolor{pole}{HTML}{FFF096}
\definecolor{traffic-sign}{HTML}{ff4b00}
\definecolor{fog}{HTML}{E60000}
\definecolor{unlabeled}{HTML}{000000}

\definecolor{SS-background}{HTML}{000000}
\definecolor{SS-car}{HTML}{6DA3FD}
\definecolor{SS-spray}{HTML}{E60000}

\definecolor{bin-correct}{HTML}{a9a9a9}
\definecolor{bin-wrong}{HTML}{ad1236}

\title{\LARGE \bf
 Label-Efficient Semantic Segmentation of LiDAR Point Clouds \\ in Adverse Weather Conditions
}

\author{Aldi Piroli$^{1}$, Vinzenz Dallabetta$^{2}$, Johannes Kopp$^{1}$, Marc Walessa$^{2}$, \\ Daniel Meissner$^{2}$, Klaus Dietmayer$^{1}$%
\thanks{$^{1}$ Institute of Measurement, Control, and Microtechnology, Ulm University, Germany {\tt\small \{firstname.lastname\}@uni-ulm.de}}
\thanks{$^{2}$ BMW~AG, Petuelring 130, 80809~Munich,~Germany {\tt\small \{vinzenz.dallabetta, marc.walessa\}@bmw.de} and {\tt\small daniel.da.meissner@bmwgroup.com}}%
}

\newcommand\copyrighttext{%
	\footnotesize \copyright\,2024 IEEE. Personal use of this material is permitted. Permission from IEEE must be obtained for all other uses, in any current or future media, including reprinting/republishing this material for advertising or promotional purposes, creating new collective works, for resale or redistribution to servers or lists, or reuse of any copyrighted component of this work in other works.}%
\newcommand\copyrightnotice{%
	\begin{tikzpicture}[remember picture,overlay]%
	\node[anchor=south,yshift=10pt] at (current page.south) {\fbox{\parbox{\dimexpr\textwidth-2cm}{\copyrighttext}}};%
	\end{tikzpicture}%
	\vspace{-10pt}%
}

\begin{document}

\maketitle
\copyrightnotice
\thispagestyle{empty}
\pagestyle{empty}

\begin{abstract}
Adverse weather conditions can severely affect the performance of LiDAR sensors by introducing unwanted noise in the measurements. 
Therefore, differentiating between noise and valid points is crucial for the reliable use of these sensors.
Current approaches for detecting adverse weather points require large amounts of labeled data, which can be difficult and expensive to obtain.
This paper proposes a label-efficient approach to segment LiDAR point clouds in adverse weather. 
We develop a framework that uses few-shot semantic segmentation to learn to segment adverse weather points from only a few labeled examples.
Then, we use a semi-supervised learning approach to generate pseudo-labels for unlabelled point clouds, significantly increasing the amount of training data without requiring any additional labeling. 
We also integrate good weather data in our training pipeline, allowing for high performance in both good and adverse weather conditions.
Results on real and synthetic datasets show that our method performs well in detecting snow, fog, and spray. 
Furthermore, we achieve competitive performance against fully supervised methods while using only a fraction of labeled data.
\end{abstract}

\section{Introduction} 
\label{sec:introduction}

LiDAR sensors are used in a wide range of robotic applications. 
Compared to cameras and radars, they provide rich and accurate depth measurements in bright and dark lighting conditions.
However, adverse weather effects like rain, snow, and fog severely degrade their performance. 
These common occurrences mainly impact LiDAR sensors in two ways. 
Firstly, due to the scattering effect, the measuring signals can be attenuated or deflected, resulting in less precise and dense measurements. 
Secondly, the water particles in fog, snow, or spray can cause partial or total signal reflection, introducing unwanted noise in the measurements. 
This last effect is particularly problematic for autonomous vehicle applications.
When driving in adverse weather conditions, a vehicle equipped with a LiDAR sensor is faced with the task of distinguishing whether a point is noise or not. 
Depending on the scenario, this distinction can be of critical importance to ensure the safe and reliable operation of the vehicle.

Semantic segmentation of LiDAR point clouds aims to assign a specific class to each point.
Despite the recent popularity of this topic, few approaches directly address the detection of adverse weather points.
CNN-based methods like~\cite{heinzler2020cnn, seppanen20224denoisenet} use standard supervised learning to segment adverse weather. 
Outlier-based methods like~\cite{piroli2023energy} rely on a scoring function to determine if a point is noise or not. 
While these approaches show promising performance, they require large amounts of labeled data to work, which can often be expensive and time-consuming to obtain.
In adverse weather conditions, the labeling task is further complicated due to artifacts introduced by these effects.
As a result, it can take an experienced human annotator several hours to label a single point cloud~\cite{xiao20233d}.

\begin{figure}[t!]
    \centering
        \includegraphics[width=\columnwidth]{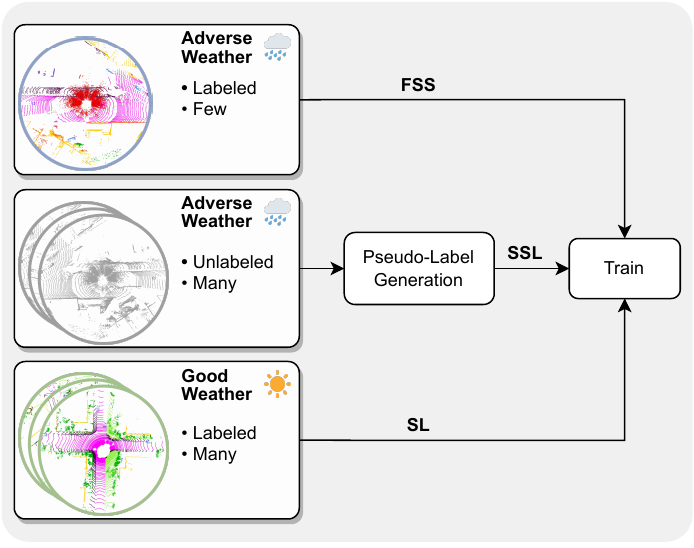}
    \caption{    
    Our proposed method employs three data sources to learn to segment LiDAR scans in adverse weather. 
    First, we use labeled point clouds recorded in adverse weather.
    As this type of data is expensive and time-consuming to obtain, we use only a few of them.
    Second, we use unlabeled data in adverse weather, which is cheap and easy to obtain.
    Third, we use labeled data in good weather conditions, which can be found in many publicly available datasets.
    We train using a combination of few-shot semantic segmentation (FSS), semi-supervised learning (SSL), and supervised learning (SL).
    }
    \label{Fig:teaser}
\end{figure}
To overcome the limitations mentioned above, we propose a label-efficient semantic segmentation method for LiDAR point clouds in adverse weather conditions.
The main idea is to use only a few annotated data in adverse weather in combination with unlabelled data. 
More specifically, we apply few-shot semantic segmentation (FSS) to first learn from a few labeled samples. 
Then, we generate pseudo-labels for the unlabeled data and use semi-supervised learning (SSL) to learn from them. 
This approach is highly efficient since data labeling is the greatest bottleneck for semantic segmentation. 
In contrast, collecting unlabelled data is cheaper and less time-consuming. 
Additionally, we integrate good weather data in our training pipeline using supervised learning (SL), allowing us to achieve high performance in both good and adverse weather conditions.
This provides a framework for the training of different FSS methods combined with a simple and yet effective SSL approach applied to the specific task of adverse weather detection. 
An overview of our data and training pipeline is shown in Fig.~\ref{Fig:teaser}.
We evaluate our approach on the SemanticSpray~\cite{piroli2023energy}, WADS~\cite{kurup2021dsor} and SemanticKITTI-fog~\cite{kong2023robo3d} datasets.
The results show that our framework can be used in combination with several different FSS approaches.
Furthermore, although our method uses only a few labeled data (i.e., $1$, $5$, $10$ scans), it achieves comparable performance to fully supervised methods in adverse weather segmentation.

In summary, our main contributions are:
\begin{itemize}
\item We propose a label-efficient method for adverse weather segmentation of LiDAR point clouds. 
\item Our framework significantly improves the performance of current LiDAR-based FSS methods in both good and adverse weather conditions.
\item  We achieve competitive performance with fully supervised approaches while using several thousand fewer labels. 
\item We evaluate our approach on both real and synthetic datasets, showing that it can generalize well to effects like snow, fog, and spray. 
\end{itemize}

\section{Related Work} 
\label{sec:related_work}
\subsection{Semantic Segmentation of LiDAR Point Clouds}
Many approaches have been proposed to segment LiDAR point clouds.
PointNet++~\cite{qi2017pointnet++} directly processes the input point cloud without any intermediate representation.
MinkowskiNet~\cite{choy20194d} first divides the point cloud in regular voxels and then uses sparse convolutions to perform segmentation.
Recently, hybrid approaches like SPVCNN~\cite{tang2020searching} have been developed to combine point and voxel processing. 

Although most methods for semantic segmentation work well in good weather, they struggle when dealing with adverse weather conditions~\cite{kong2023robo3d}. %
Only a few approaches have been proposed to detect adverse weather points in 3D point clouds~\cite{dreissig2023survey,piroli2022detection,piroli2023towards}. 
Non-learning-based methods, such as Kurup et al.~\cite{kurup2021dsor}, filter noise points using statistical information like the number of neighbors.
Their performance is usually inferior to learning-based approaches~\cite{piroli2023energy}, and they tend to require a great deal of manual fine-tuning.  
Deep learning methods like Heinzler et al.~\cite{heinzler2020cnn} use a CNN-based approach to detect fog and rainfall in LiDAR point clouds. 
Piroli et al.~\cite{piroli2023energy} instead propose an outlier detection framework for identifying adverse weather points like snow, fog, and rain spray.
Sepp{\"a}nen et al.~\cite{seppanen20224denoisenet} use successive scans to include temporal information to segment snow points. 
Although these methods have shown promising results in detecting adverse weather points, they all require a large number of manually annotated data.

\subsection{Few-shot Semantic Segmentation}
Few-shot semantic segmentation (FSS) aims to reduce the required number of labeled data for semantic segmentation.
A large body of work exists for the image domain~\cite{catalano2023few}.

In the 3D point cloud domain, the FSS task remains largely unexplored.
Mei et al.~\cite{10160674} propose a method for FSS on outdoor LiDAR point clouds. 
They use a novel loss function to deal with the background ambiguity of classes that can arise during FSS training. 
Additionally, they apply other FSS methods developed for imagery to 3D point clouds.
In particular, they study the use of knowledge distillation combined with fine-tuning~\cite{li2017learning} and fine-tuning combined with contrastive learning~\cite{myers2021generalized}.
In this work, we use~\cite{10160674, li2017learning, myers2021generalized} as FSS baselines for our method.

Recently, Liu et al.~\cite{liu2022less} have proposed a label-efficient approach for the segmentation of LiDAR point clouds. 
The method is based on the assumption that once the ground points have been detected, most of the objects can be easily clustered into well-identifiable point clouds.
This assumption is often not valid for 3D point clouds in adverse weather conditions, where unwanted noise like spray, snow, and fog is often scattered near objects, making the separation difficult. 
In addition, when the ground is wet, many ground points are not detected due to the scattering effect, making the required clustering even more challenging.
Finally, although their approach achieves good segmentation performance with only a small number of labeled data, the required number is still orders of magnitude greater than in an FSS setting.

\subsection{Semi-supervised Learning}
Semi-supervised learning (SSL) aims to improve the performance of a model by using unlabeled data.
Many works have been proposed for the image domain~\cite{yang2022survey}.
For example, consistency regularisation optimizes the model to return the same features for different representations of the same input.
Other methods like Li et al.~\cite{li2019learning} use pseudo-labels during training to improve performance in few-shot learning scenarios.
Wang et al.~\cite{wang2019symmetric} propose the symmetric cross-entropy, which allows robust learning with noisy labels.
Recently, Xie et al.~\cite{xie2020pointcontrast} have proposed an SSL method for pre-training 3D point cloud processing networks.
Their approach adapts the InfoNCE loss~\cite{oord2018representation} to the 3D domain by encouraging the same feature representation for different views of the same point cloud. 
Li et al.~\cite{jiang2021guided} refine the approach using a dual branch network that includes supervised training to refine the SSL loss function.
Our method uses SSL to learn from unlabeled data by generating pseudo-labels and then employs the symmetric cross entropy~\cite{wang2019symmetric} during optimization.

\section{Method}
\label{sec:method}
\begin{figure*}[t!]
    \centering
    \includegraphics[width=0.90\textwidth]{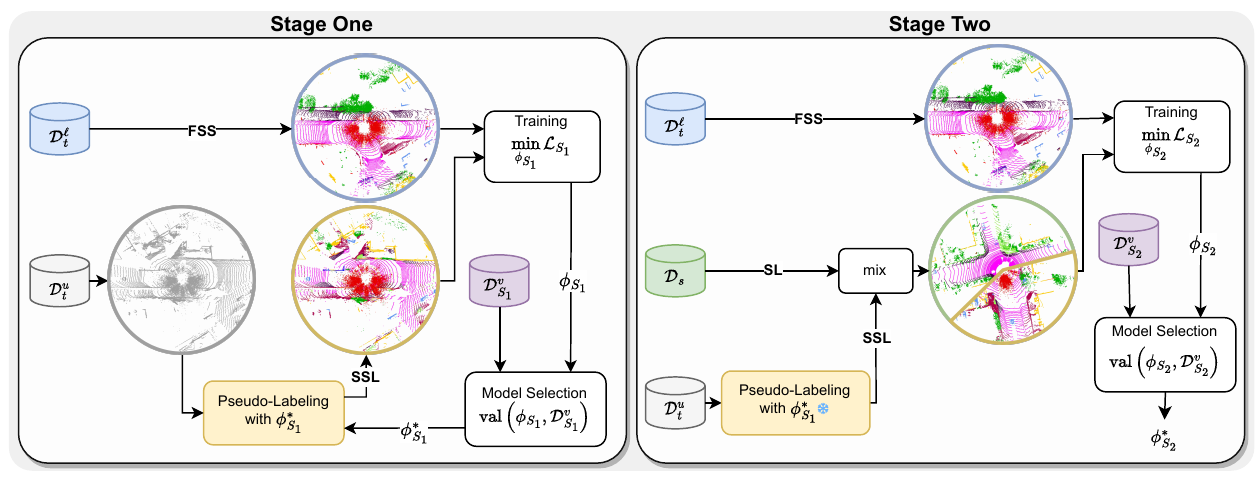}
    \caption{ 
    Overview of our method.
    \textbf{Stage One}: we aim to train a model $\phi_{S_1}$ that performs well in adverse weather scenes.
    We use few-shot semantic segmentation (FSS) to learn from the labeled dataset $\mathcal{D}_t^\ell$ (adverse weather).
    We iteratively select the best-performing model $\phi_{S_1}^*$ by validating the performance of $\phi_{S_1}$ on the pseudo-validation set $\mathcal{D}_{S_1}^v$.
    To obtain additional training data without increasing the labeling effort, we generate pseudo-labels for the unlabelled target dataset $\mathcal{D}_t^u$ using $\phi_{S_1}^*$ and train with them in a semi-supervised learning (SSL) fashion.
    \textbf{Stage Two}: we aim to train a new model $\phi_{S_2}$ that performs well in both good and adverse weather. 
    We use FSS training like in the previous stage. 
    Additionally, we combine pseudo-labels generated from the resulting model $\phi_{S_1}^*$ of Stage One with labeled data sampled from $\mathcal{D}_s$ (good weather) using polar data mixing and train with them using a combination of supervised-learning (SL) and SSL.
    We select the best-performing model $\phi_{S_2}^*$ by validating $\phi_{S_2}$ on the pseudo-validation set $\mathcal{D}_{S_2}^v$.
    }
    \label{Fig:method}
\end{figure*}
We aim to train a model that can semantically segment LiDAR point clouds in adverse weather using only $K$ labeled scans.
Our method consists of two stages.
In the first stage, we train a model using FSS on $K$ labeled scans.
Additionally, we use this model to generate pseudo-labels for unlabelled data and learn from them using SSL.
In the second stage, we combine FSS, SSL, and SL to train a model on both good and adverse weather scenes. 
An overview of our approach is shown in Fig.~\ref{Fig:method}.

\subsection{Problem Setting}
We assume to have a source dataset $\mathcal{D}_s$ that was recorded in good weather conditions. 
It includes many different scenes containing base classes $C_b$ (e.g., vehicle, pedestrian, vegetation, $\dots$).
Each scene consists of a point cloud $\*x \in \mathbb{R} ^ {N\times F}$ with $N$ points and $F$ features, and the associated labels $\*y \in \mathbb{N} ^ {N}$.
Additionally, we have a target dataset $\mathcal{D}_t = \mathcal{D}_t^\ell \cup \mathcal{D}_t^u$ recorded in adverse weather conditions containing novel classes $C_n$ (e.g., snow, fog, spray, $\dots$).
The subset $\mathcal{D}_t^\ell$ contains only $K$ labeled scenes (shots), whereas $\mathcal{D}_t^u$ is large but does not have any labeled data.
It is assumed that $C_b \cap C_n = \emptyset$ so that the source dataset does not contain any novel classes.

\subsection{Stage Zero}
Following standard FSS approaches, we train a base model $\phi_{S_0} (\*x): \mathbb{R}^{N \times F} \rightarrow \mathbb{R}^{N \times \mid C_b \mid }$  on the source dataset $\mathcal{D}_s$.
This is done in a fully supervised manner using a loss function like the cross-entropy.
However, depending on the FSS method, additional losses may also be included.

\subsection{Stage One}
\label{sec:stage_one}
In this stage, we aim to train a model $\phi_{S_1} (\*x): \mathbb{R}^{N \times F} \rightarrow \mathbb{R}^{N \times \mid C_b \mid + \mid C_n \mid }$ which can segment point clouds in adverse weather.
We do this by training with a combination of FSS and SSL.
More precisely, we initialize the weights of $\phi_{S_1} \leftarrow \phi_{S_0}$ using the base model trained in Stage Zero (the additional $C_n$ weights of the last classification layer are randomly initialized) and train it in an FSS fashion by minimizing the loss
\begin{align}\label{eq:stage_1_FSS_loss}
    \mathcal{L}_\text{FSS} = \sum_{i=1}^{K} L_\text{FSS}\big(\phi_{S_1}(\*x_i), \*y_i\big),
\end{align}
where $L_\text{FSS}$ is defined by the FSS method of choice and $ (\*x_i, \*y_i) \in \mathcal{D}_t^{\ell}$.

In addition to the $K$ labeled scans, we also use unlabeled data during training. 
We do so by first selecting the best-performing model $\phi_{S_1}^*$ using the validation process described in Section~\ref{method:best_model_selection}.
Then, we generate pseudo-labels by performing inference on the unlabeled data $\mathcal{D}_t^u$.
By iteratively updating the best model through this stage, we can continuously improve the quality of the generated labels. 
This allows us to learn from a much larger data pool without increasing the labeling effort. 
To reduce the effect of noise in the pseudo-labels, we use the symmetric cross entropy loss~\cite{wang2019symmetric}
\begin{align}\label{eq:stage_1_SSL_loss}
    \mathcal{L}_\text{SSL} = \sum_{i=1}^{K} L_\text{SCE}\big(\phi_{S_1}(\*x_i), \*y_i\big),
\end{align}
where $\*x_i \in \mathcal{D}_t^{u}$ and $\*y_i = \text{argmax}\big(\phi_{S_1}^*(\*x_i)\big)$.

The total loss function used in Stage One is defined as
\begin{align}\label{eq:stage_1_total_loss}
    \mathcal{L}_{\text{S}_1} = \mathcal{L}_\text{FSS} + \omega_0 \mathcal{L}_\text{SSL},
\end{align}
where $\omega_0 \in \mathbb{R}$ is a warmup and weight parameter.
To avoid training with extremely noisy pseudo-labels generated in early training stages, we set $\omega_0 = 0$ if the mean intersection-over-union (mIoU) of $\phi_{S_1}^*$ on the pseudo-validation set $\mathcal{D}_{S_1}^v$ (described Section~\ref{Section:PseudoVal}) is less than a threshold $\gamma \in \mathbb{R}$.

\subsection{Stage Two}
In this stage, we aim to train a model $\phi_{S_2} (\*x): \mathbb{R}^{N \times F} \rightarrow \mathbb{R}^{N \times \mid C_b \mid + \mid C_n \mid }$ which performs well on both good and adverse weather data.
Similar to Stage One, we initialize the weights of $\phi_{S_2} \leftarrow \phi_{S_0}$ from the base model trained in Stage Zero.
We avoid using the weights of $\phi_{S_1}^*$ for initialization as it is only trained to perform well on the target data set, and its performance on the base classes might then be lower than that of $\phi_{S_0}$.
We employ the same FSS loss function as in~\eqref{eq:stage_1_FSS_loss} to learn from the $K$ adverse weather labeled data
\begin{align}\label{eq:stage_2_FSS_loss}
    \mathcal{L}_\text{FSS} = \sum_{i=1}^{K} L_\text{FSS}\big(\phi_{S_2}(\*x_i), \*y_i\big),
\end{align}
where $(\*x_i, \*y_i) \in \mathcal{D}_t^{\ell}$.
Additionally, we use the knowledge distillation loss proposed in~\cite{hinton2015distilling} to help retain the performance of the base model $\phi_{S_0}$ on the base classes $C_b$
\begin{align}\label{eq:stage_2_KD_sup}
    \mathcal{L}_\text{KD} = \sum_{i=1}^{K } \sum_{j\in C_b} \text{KD}\big(\phi_{S_2}^j(\*x_i),\phi_{S_0}^j(\*x_i)\big),
\end{align}
where  $\*x_i \in \mathcal{D}_t^{\ell}$ and $\phi_{S_0}^j(\*x_i)$, $\phi_{S_2}^j(\*x_i)$ represent the $j$-th output of the models.

To further improve performance in both good and adverse weather, we use the  $\mathcal{D}_s$ and $\mathcal{D}_t^u$ datasets.
We randomly sample unlabeled point clouds from $\mathcal{D}_t^u$ and generate the corresponding pseudo-labeled using $\phi_{S_1}^*$ so that $\mathcal{X}^u = \{(\*x_i, \text{argmax}\left(\phi_{S_1}^*(\*x_i)\right) \mid \*x_i \in \mathcal{D}_t^u\}_{i=1}^{K}$.
We use $\phi_{S_1}^*$, which is the resulting model from Stage One for pseudo-label generation, as it is specifically trained to achieve high performance on the target dataset. 
We also randomly sample labeled scans from the source dataset $\mathcal{D}_s$ so that $\mathcal{X}^\ell = \{(\*x_i, \*y_i) \mid (\*x_i, \*y_i) \in \mathcal{D}_s\}_{i=1}^{K}$.
Then, we combine the two sets using a polar data mixing augmentation similar to~\cite{xiao2022polarmix}, such that $\mathcal{X}_\text{mix} = \text{mix}(\mathcal{X}^u, \mathcal{X}^\ell)$.
As loss function, we use both the symmetric cross entropy~\cite{wang2019symmetric} and knowledge distillation~\cite{hinton2015distilling} loss
\begin{align}\label{eq:stage_2_mixup}
    \mathcal{L}_\text{mix} = &\sum_{i=1}^{K}  L_\text{SCE}\big( \phi_{S_2}(\*x_i), \*y_i\big) + \nonumber \\
    & \sum_{i=1}^{K} \sum_{j\in C_b} \text{KD}\big(\phi_{S_2}^j(\*x_i), \phi_{S_0}^j(\*x_i)\big),
\end{align}
where $(\*x_i, \*y_i) \in \mathcal{X}_\text{mix}$.
As $\mathcal{X}_\text{mix}$ contains a combination of ground truth and pseudo-labels, the loss~\eqref{eq:stage_2_mixup} combines both SL and SSL.
The total loss function for Stage Two is defined as
\begin{align}\label{eq:stage_2_total_loss}
    \mathcal{L}_{\text{S}_2} = \mathcal{L}_\text{FSS} + \omega_1 \mathcal{L}_\text{KD} + \omega_2 \mathcal{L}_\text{mix},
\end{align}
where $\omega_1, \omega_2 \in \mathbb{R}$ are weighting parameters.

\subsection{Pseudo-Validation Set}
\label{Section:PseudoVal}
\begin{figure}[t!]
    \centering
    \includegraphics[width=0.9\columnwidth]{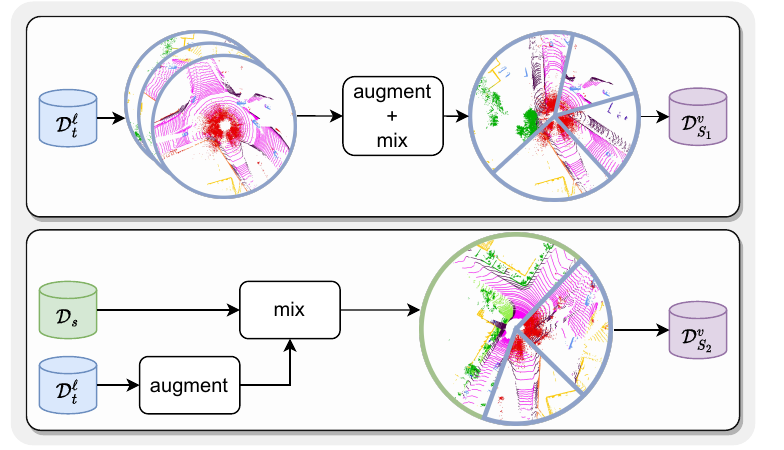}
    \caption{ Overview of the pseudo-validation sets generation.
    Top: to generate the pseudo-validation set for Stage One $\mathcal{D}_{S_1}^v$ we use scans from the $K$ labeled data in $\mathcal{D}_t^\ell$, augment them, and then use polar mixing to generate new point clouds.
    Bottom: for the pseudo-validation set for Stage Two $\mathcal{D}_{S_2}^v$, we additionally use data in good weather sampled from $\mathcal{D}_s$.
    }
    \label{Fig:pseudo_val}
\end{figure}

In an FSS setting, given the limited number of labeled data, it is unreasonable to assume that a large validation set is available.
However, validating which checkpoint is the best is crucial, given the risk of overfitting the small number of training data.
We propose to solve this problem by generating a pseudo-validation dataset using only the $K$ labeled samples in $\mathcal{D}_t^\ell$.
Since the scans in $\mathcal{D}_t^\ell$ are also used during training, we apply to each point cloud a series of augmentations like random axis flip, rotation, size scaling, and intensity scaling.
To further increase data variance, we add random noise to the position $(x,y,z)$ of the adverse weather points.
For Stage One, we construct a validation set $\mathcal{D}_{S_1}^v$ by randomly mixing polar sections from the $K$ augmented point clouds sampled from $\mathcal{D}_t^\ell$.
For Stage Two, we augment point clouds from $\mathcal{D}_t^\ell$ and randomly mix them with $\mathcal{D}_s$ and obtain the validation set $\mathcal{D}_{S_2}^v$.
An example of scans from both $\mathcal{D}_{S_1}^v$ and $\mathcal{D}_{S_2}^v$ is given in Fig.~\ref{Fig:pseudo_val}.

\subsection{Best Model Selection}
\label{method:best_model_selection}
To select the best-performing model $\phi_{S_1}^*$ of Stage One, we validate the performance of $\phi_{S_1}$ in terms on mIoU every $M$ epochs on the pseudo-validation set $\mathcal{D}_{S_1}^v$, such that  $\text{val}(\phi_{S_1}, \mathcal{D}_{S_1}^v) \Rightarrow \phi_{S_1}^*$.
Similarly, for Stage Two, we evaluate every $M$ epochs using $\text{val}(\phi_{S_2}, \mathcal{D}_{S_2}^v) \Rightarrow \phi_{S_2}^*$ .

\section{Evaluation} 
\label{sec:experiments}

\subsection{Experiment Setup}
\textbf{Datasets.}
We evaluate our approach on three different adverse weather datasets.
The SemanticSpray dataset~\cite{piroli2023energy} contains highway-like scenarios of vehicles traveling at different speeds on wet road surfaces, generating spray as a result.
The dataset contains $7898$ scans for training and $8667$ scans for testing.
The WADS dataset~\cite{kurup2021dsor} was recorded in an urban environment while in snowy conditions.
Following~\cite{piroli2023energy}, we split the dataset into $1011$ scans for training and $918$ for testing.
To test the performance on foggy conditions, we use the SemanticKITTI-fog (moderate)~\cite{kong2023robo3d} dataset, which was generated from the SemanticKITTI dataset~\cite{behley2019semantickitti} using a physically valid fog model simulation~\cite{hahner2021fog}.
We split the dataset evenly so that both the training and test sets have $1831$ scans.
As source datasets $\mathcal{D}_s$ in good weather conditions, we use the SemanticKITTI~\cite{behley2019semantickitti} and nuScenes datasets~\cite{caesar2020nuscenes}, adopting the official train and validation splits.

\textbf{FSS Baselines.}
As baselines for FSS, we use three state-of-the-art approaches for outdoor LiDAR point clouds (FSSAD~\cite{10160674}, GFSS~\cite{myers2021generalized} and LwF~\cite{li2017learning}).
For FSSAD~\cite{10160674}, we train $\phi_{S_0}$ using the cross-entropy and Lovász-softmax functions described by the authors. 
For GFSS~\cite{myers2021generalized}, we train $\phi_{S_0}$ using the cross-entropy and triplet loss regularization.
For the latter, we use a maximum of $50$ anchors per training step and set the margin parameter to $1$.
Similar to what is reported in~\cite{10160674}, we find that the best performance for GFSS is obtained when fine-tuning both the backbone and segmentation head.
Finally, for LwF~\cite{li2017learning} we train $\phi_{S_0}$ using the cross-entropy loss function.
We use SemanticKITTI as the good weather source dataset for the target datasets WADS and SemanticKITTI-fog. 
For SemanticSpray, we use nuScenes as good weather data instead.
These combinations are chosen so that the source and target datasets share a similar LiDAR sensor ($64$~layers for SemanticKITTI and $32$~layers for nuScenes) so that the domain gap is reduced.

\textbf{SSL Baselines.}
We also compare with other SSL approaches, namely PointContrast~\cite{xie2020pointcontrast} and GuidedPointContrast~\cite{jiang2021guided}.
For PointContrast~\cite{xie2020pointcontrast}, we empirically set the temperature parameter to $\tau = 100$.
For GuidedPointContrast~\cite{jiang2021guided}, we follow the author's training pipeline and pretrain the fully supervised branch for $500$ epochs and then set $\tau=100$ and confidence score threshold to $0.75$ IoU in the SSL branch.
For both PointContrast~\cite{xie2020pointcontrast} and GuidedPointContrast~\cite{jiang2021guided}, we adopt a learning strategy that includes the SSL loss combined with the FSS loss of LwF, such that the total loss for a single optimization step is $\mathcal{L}_\text{tot} = \mathcal{L}_\text{SSL} + \mathcal{L}_\text{FSS}$. 
We highlight that although the SSL baselines are not used for pretraining, the base model $\phi_{S_0}$ is still pretrained following the LwF FSS strategy. 
When comparing our proposed method to the other SSL baselines, we use Stage Zero for the base model pertaining and then Stage One for the combined FSS and SSL.

\textbf{Implementation Details.}
We use SPVCNN~\cite{tang2020searching} as the semantic segmentation network for all methods.
In all experiments, we train with SGD using learning rate $1\times10^{-3}$, weight decay $1\times10^{-4}$ and momentum $0.9$.
As a reference, we also train SPVCNN in a fully supervised manner on the target datasets ($15$ epochs on SemanticKITTI-fog, $30$ epochs for SemanticSpray, and $100$ epochs for WADS).
In Stage Zero of our method, we train for $15$ epochs when using the SemanticKITTI dataset and for $30$ epochs when using nuScenes. 
For Stage One and Two, we train for $\si{2}{k}$  and $\si{3}{k}$ epochs respectively.
Unless otherwise specified, we set the weight parameters $\omega_0=\omega_1=\omega_2 = 0.5$ and the mIoU threshold for Stage One $\gamma = 0.75$.
The pseudo-validation sets sizes $\mid\mathcal{D}_{S_1}^v\mid, \mid\mathcal{D}_{S_2}^v\mid$ are equal to $500$.
We perform model selection every $M=50$ epochs.
As augmentation for the pseudo-validation set, we use random axis flip, random rotation with angle range $[-\frac{\pi}{4}, \frac{\pi}{4}]$, random size scaling with range $[0.9,1.1]$ and intensity scaling with range $[0.9,1.0]$.
We add noise to adverse weather points sampled from a normal distribution $\mathcal{N}(0, 0.3)$. 
The polar mixing operation used across our method takes two scans and combines them by randomly selecting an angle $\theta \in (0, 2\pi)$ from the first and inserting it in the second one. 
Similarly, it combines more than two scans by randomly selecting polar sections from all scans to form a full $360^{\circ}$ scan.
To ensure a fair comparison, we train using the same random seed for each different few-shot combination so that the same scenes are used for both the baseline and the proposed method.

\subsection{Results}
\begin{table}[t!]
    \centering
    \caption{Results on the SemanticSpray dataset using $K$ shots. 
    The first row shows the results when training with $100\%$ of the data.
    We report the IoU for each class and the resulting mIoU. 
    We also include the mIoU~$\mathcal{D}_s$ on the source dataset (nuScenes). 
    The superscript~$^*$ indicates the novel classes $C_n$.
    }
    \resizebox{0.9\columnwidth}{!}{%

        \begin{tabular}{l|l|ccc|c|c}
            \hline
            $K$                   & Method                    & \rotatebox{90}{\textit{vehicle}} & \rotatebox{90}{\textit{background}} & \rotatebox{90}{\textit{spray$^*$}} & \rotatebox{90}{mIoU} & \rotatebox{90}{mIoU~$\mathcal{D}_s$} \\  \hline
                                  & \textit{fully supervised} & 95.24                        & 99.82                           & 88.40                          & 94.49                & 71.86                                \\ \hline
            \multirow{6}{*}{$1$}  & FSSAD                     & 32.60                        & 99.16                           & 53.37                          & 61.71                & 4.58                                 \\
                                  & FSSAD + ours              & \textbf{93.92}               & \textbf{99.64}                  & \textbf{76.14}                 & \textbf{89.90}       & \textbf{60.81}                       \\ %
                                  & GFSS                      & 33.24                        & 98.89                           & 41.38                          & 57.84                & \textbf{8.37}                        \\
                                  & GFSS + ours               & \textbf{89.36}               & \textbf{99.25}                  & \textbf{42.57}                 & \textbf{77.06}       & 6.71                                 \\ %
                                  & LwF                       & 12.73                        & 99.00                           & 43.65                          & 51.79                & 11.01                                \\
                                  & LwF + ours                & \textbf{56.22}               & \textbf{99.68}                  & \textbf{52.05}                 & \textbf{69.32}       & \textbf{62.35}                       \\ \hline
            \multirow{6}{*}{$5$}  & FSSAD                     & 91.14                        & 99.72                           & 80.67                          & 90.51                & 6.35                                 \\
                                  & FSSAD + ours              & \textbf{94.52}               & \textbf{99.77}                  & \textbf{85.16}                 & \textbf{93.15}       & \textbf{64.03}                       \\ %
                                  & GFSS                      & 74.55                        & 99.67                           & 69.47                          & 81.23                & 4.91                                 \\
                                  & GFSS + ours               & \textbf{93.85}               & 99.67                           & \textbf{77.00}                 & \textbf{90.17}       & \textbf{49.02}                       \\ %
                                  & LwF                       & 66.45                        & 99.67                           & 73.85                          & 79.99                & 8.41                                 \\
                                  & LwF + ours                & \textbf{94.10}               & \textbf{99.72}                  & \textbf{80.93}                 & \textbf{91.58}       & \textbf{62.89}                       \\ \hline
            \multirow{6}{*}{$10$} & FSSAD                     & 93.61                        & 99.75                           & 83.95                          & 92.44                & 5.05                                 \\
                                  & FSSAD + ours              & \textbf{94.23}               & \textbf{99.77}                  & \textbf{86.02}                 & \textbf{93.34}       & \textbf{63.04}                       \\ %
                                  & GFSS                      & 80.27                        & \textbf{99.69}                  & 72.72                          & 84.23                & 4.43                                 \\
                                  & GFSS + ours               & \textbf{93.35}               & 99.66                           & \textbf{76.27}                 & \textbf{89.76}       & \textbf{46.88}                       \\ %
                                  & LwF                       & 83.89                        & 99.72                           & 81.25                          & 88.29                & 7.86                                 \\
                                  & LwF + ours                & \textbf{95.21}               & \textbf{99.77}                  & \textbf{84.91}                 & \textbf{93.30}       & \textbf{63.37}                       \\ \hline
        \end{tabular}

    }
    \label{table:FSS_SemanticSpray}
\end{table}

\begin{table*}[t!]
    \centering
    \caption{Results on the WADS dataset using $K$ shots. 
    The first row shows the results when training with $100\%$ of the data.
    We report the IoU for each class and the resulting mIoU. 
    We also include the mIoU~$\mathcal{D}_s$ on the source dataset (SemanticKITTI). 
    The superscript~$^*$ indicates the novel classes $C_n$.
    }
    \resizebox{0.95\textwidth}{!}{%

        \begin{tabular}{c|l|cccccccccccccccc|c|c}
            \hline
            $K$                   & Method                    & \rotatebox{90}{\textit{car}} & \rotatebox{90}{\textit{truck}} & \rotatebox{90}{\textit{other-vehicle}} & \rotatebox{90}{\textit{person}} & \rotatebox{90}{\textit{road}} & \rotatebox{90}{\textit{parking}} & \rotatebox{90}{\textit{sidewalk}} & \rotatebox{90}{\textit{other-ground}} & \rotatebox{90}{\textit{building}} & \rotatebox{90}{\textit{fence}} & \rotatebox{90}{\textit{vegetation}} & \rotatebox{90}{\textit{trunk}} & \rotatebox{90}{\textit{terrain}} & \rotatebox{90}{\textit{pole}} & \rotatebox{90}{\textit{traffic-sign}} & \rotatebox{90}{\textit{snow$^*$}} & \rotatebox{90}{mIoU} & \rotatebox{90}{$\text{mIoU}~\mathcal{D}_s$} \\ \hline
                                  & \textit{fully supervised} & 63.30                        & 3.97                           & 2.43                                   & 0.01                            & 60.46                         & 2.20                             & 21.76                             & 3.24                                  & 70.80                             & 16.47                          & 57.60                               & 0.00                           & 0.00                             & 20.64                         & 15.28                                 & 76.44                              & 25.91                & 64.14                                       \\ \hline
            \multirow{6}{*}{$1$}  & FSSAD                     & 50.13                        & 2.75                           & \textbf{0.57}                          & 0.00                            & \textbf{57.68}                & \textbf{2.12}                    & 3.69                              & \textbf{0.45}                         & 74.58                             & 16.10                          & 8.51                                & \textbf{5.91}                  & \textbf{0.46}                    & \textbf{23.75}                & \textbf{13.25}                        & 61.83                              & 20.11                & 29.17                                       \\
                                  & FSSAD +  ours             & \textbf{58.71}               & \textbf{12.77}                 & 0.08                                   & 0.00                            & 53.58                         & 0.35                             & \textbf{8.50}                     & 0.00                                  & \textbf{79.29}                    & \textbf{21.51}                 & \textbf{25.88}                      & 4.43                           & 0.02                             & 22.98                         & 2.53                                  & \textbf{70.78}                     & \textbf{22.59}       & \textbf{51.42}                              \\ %
                                  & GFSS                      & 48.22                        & 10.34                          & \textbf{0.36}                          & 0.00                            & 57.52                         & \textbf{2.89}                    & 0.71                              & \textbf{0.13}                         & 69.61                             & 5.84                           & 3.57                                & \textbf{4.05}                  & 0.48                             & \textbf{19.45}                & \textbf{16.70}                        & 53.81                              & 18.36                & 23.63                                       \\
                                  & GFSS +  ours              & \textbf{61.07}               & \textbf{11.94}                 & 0.01                                   & 0.00                            & \textbf{58.00}                & 0.61                             & \textbf{5.89}                     & 0.05                                  & \textbf{78.52}                    & \textbf{20.54}                 & \textbf{36.74}                      & 1.15                           & \textbf{0.65}                    & \textbf{4.13}                 & 13.79                                 & \textbf{77.84}                     & \textbf{23.18}       & \textbf{43.96}                              \\ %
                                  & LwF                       & 39.68                        & 10.47                          & 0.63                                   & \textbf{4.30}                   & 56.51                         & \textbf{1.04}                    & 5.97                              & 1.66                                  & 75.55                             & 16.94                          & 13.46                               & \textbf{3.26}                  & \textbf{0.26}                    & 20.59                         & 17.36                                 & 62.00                              & 20.60                & 35.29                                       \\
                                  & LwF +  ours               & \textbf{61.31}               & \textbf{12.37}                 & \textbf{0.66}                          & 4.13                            & \textbf{58.70}                & 0.72                             & \textbf{7.59}                     & \textbf{2.01}                         & \textbf{78.11}                    & \textbf{22.49}                 & \textbf{28.45}                      & 2.13                           & 0.15                             & 17.52                         & \textbf{19.29}                        & \textbf{73.20}                     & \textbf{24.30}       & \textbf{52.27}                              \\ \hline
            \multirow{6}{*}{$5$}  & FSSAD                     & 65.74                        & 5.30                           & \textbf{2.07}                          & 0.54                            & 56.31                         & 0.18                             & 4.36                              & \textbf{12.98}                        & 72.99                             & 6.49                           & 53.27                               & 0.27                           & \textbf{0.22}                    & 16.85                         & 9.99                                  & 76.29                              & 23.99                & 27.11                                       \\
                                  & FSSAD +  ours             & \textbf{72.00}               & \textbf{40.40}                 & 1.50                                   & \textbf{1.73}                   & \textbf{58.03}                & \textbf{0.22}                    & \textbf{9.84}                     & 7.54                                  & \textbf{79.84}                    & \textbf{16.63}                 & \textbf{57.79}                      & \textbf{3.05}                  & 0.10                             & \textbf{20.25}                & \textbf{17.49}                        & \textbf{79.36}                     & \textbf{29.11}       & \textbf{54.51}                              \\ %
                                  & GFSS                      & 47.40                        & 8.30                           & \textbf{0.99}                          & \textbf{0.53}                   & 52.89                         & \textbf{0.48}                    & 2.51                              & \textbf{8.51}                         & 67.20                             & 2.47                           & 47.04                               & 0.34                           & \textbf{0.10}                    & \textbf{15.01}                & 9.21                                  & 69.24                              & 20.76                & 20.66                                       \\
                                  & GFSS +  ours              & \textbf{68.34}               & \textbf{8.64}                  & 0.39                                   & 0.00                            & \textbf{58.82}                & 0.20                             & \textbf{4.61}                     & 5.84                                  & \textbf{77.47}                    & \textbf{14.49}                 & \textbf{54.93}                      & \textbf{0.55}                  & 0.03                             & 8.21                          & \textbf{13.82}                        & \textbf{73.40}                     & \textbf{24.36}       & \textbf{42.90}                              \\ %
                                  & LwF                       & 60.42                        & 11.45                          & 1.13                                   & \textbf{4.68}                   & 56.52                         & \textbf{0.60}                    & 7.33                              & \textbf{6.57}                         & 75.65                             & 12.50                          & 47.12                               & \textbf{1.95}                  & \textbf{0.17}                    & 17.55                         & 11.69                                 & 76.25                              & 24.47                & 37.84                                       \\
                                  & LwF +  ours               & \textbf{67.50}               & \textbf{13.82}                 & \textbf{1.99}                          & 2.98                            & \textbf{58.47}                & 0.19                             & \textbf{7.95}                     & 5.10                                  & \textbf{80.33}                    & \textbf{19.30}                 & \textbf{52.50}                      & 0.79                           & 0.12                             & \textbf{21.23}                & \textbf{18.06}                        & \textbf{79.24}                     & \textbf{26.85}       & \textbf{57.40}                              \\ \hline
            \multirow{6}{*}{$10$} & FSSAD                     & 67.20                        & 3.41                           & 1.32                                   & 0.94                            & 54.23                         & \textbf{2.42}                    & 1.22                              & \textbf{13.20}                        & 74.31                             & 7.74                           & 53.86                               & 0.19                           & \textbf{0.20}                    & \textbf{18.46}                & 13.76                                 & 75.58                              & 24.25                & 23.78                                       \\
                                  & FSSAD +  ours             & \textbf{73.39}               & \textbf{15.02}                 & \textbf{1.71}                          & \textbf{3.64}                   & \textbf{56.45}                & 0.62                             & \textbf{5.36}                     & 10.99                                 & \textbf{80.33}                    & \textbf{17.98}                 & \textbf{61.80}                      & \textbf{1.39}                  & 0.05                             & 16.61                         & \textbf{18.69}                        & \textbf{78.57}                     & \textbf{27.66}       & \textbf{59.25}                              \\ %
                                  & GFSS                      & 47.57                        & 7.49                           & 1.06                                   & 0.62                            & 52.27                         & \textbf{2.00}                    & 1.35                              & 8.37                                  & 67.56                             & 2.75                           & 46.12                               & 0.53                           & \textbf{0.08}                    & \textbf{15.69}                & 7.07                                  & 64.05                              & 20.29                & 16.93                                       \\
                                  & GFSS +  ours              & \textbf{68.20}               & \textbf{5.29}                  & \textbf{1.79}                          & \textbf{0.77}                   & \textbf{55.38}                & 0.83                             & \textbf{4.42}                     & \textbf{8.73}                         & \textbf{75.54}                    & \textbf{11.65}                 & \textbf{61.83}                      & \textbf{0.14}                  & 0.03                             & 15.27                         & \textbf{16.11}                        & \textbf{71.91}                     & \textbf{24.87}       & \textbf{49.61}                              \\ %
                                  & LwF                       & 58.87                        & 13.63                          & 0.75                                   & \textbf{5.13}                   & \textbf{56.43}                & \textbf{2.16}                    & \textbf{7.26}                     & 8.70                                  & 77.55                             & \textbf{19.31}                 & 43.80                               & 2.32                           & 0.17                             & 18.27                         & 15.00                                 & 70.49                              & 24.99                & 36.17                                       \\
                                  & LwF +  ours               & \textbf{71.24}               & \textbf{24.49}                 & \textbf{2.56}                          & 2.84                            & 55.44                         & 0.16                             & 6.48                              & \textbf{9.45}                         & \textbf{79.72}                    & 16.58                          & \textbf{46.35}                      & \textbf{4.17}                  & \textbf{0.25}                    & \textbf{24.18}                & \textbf{18.18}                        & \textbf{77.91}                     & \textbf{27.50}       & \textbf{58.49}                              \\ \hline
        \end{tabular}
    }
        \label{table:FSS_WADS}
\end{table*}

\begin{table*}[t!]
    \centering
    \caption{Results on the SemanticKITTI-fog dataset using $K$ shots. 
    The first row shows the results when training with $100\%$ of the data.
    We report the IoU for each class and the resulting mIoU. 
    We also include the mIoU~$\mathcal{D}_s$ on the source dataset (SemanticKITTI). 
    The superscript~$^*$ indicates the novel classes $C_n$. 
    }
    \resizebox{0.95\textwidth}{!}{%
        \begin{tabular}{c|l|ccccccccccccccccccc|c|c}
            \hline
            $K$                   & Method                    & \rotatebox{90}{\textit{car}} & \rotatebox{90}{\textit{bicycle}} & \rotatebox{90}{\textit{motorcycle}} & \rotatebox{90}{\textit{truck}} & \rotatebox{90}{\textit{other-vehicle}} & \rotatebox{90}{\textit{person}} & \rotatebox{90}{\textit{bicyclist}} & \rotatebox{90}{\textit{road}} & \rotatebox{90}{\textit{parking}} & \rotatebox{90}{\textit{sidewalk}} & \rotatebox{90}{\textit{other-ground}} & \rotatebox{90}{\textit{building}} & \rotatebox{90}{\textit{fence}} & \rotatebox{90}{\textit{vegetation}} & \rotatebox{90}{\textit{trunk}} & \rotatebox{90}{\textit{terrain}} & \rotatebox{90}{\textit{pole}} & \rotatebox{90}{\textit{traffic-sign}} & \rotatebox{90}{\textit{fog$^*$}} & \rotatebox{90}{mIoU} & \rotatebox{90}{mIoU~$\mathcal{D}_s$} \\ \hline
                                  & \textit{fully supervised} & 90.38                        & 42.50                            & 66.92                               & 79.69                          & 57.47                                  & 68.33                           & 82.73                              & 91.56                         & 52.22                            & 82.12                             & 55.41                                 & 90.72                             & 74.33                          & 90.22                               & 70.77                          & 80.86                            & 64.24                         & 48.02                                 & 92.28                        & 69.04                & 64.14                                \\ \hline
            \multirow{6}{*}{$1$}  & FSSAD                     & 88.93                        & \textbf{29.53}                   & 42.04                               & 63.99                          & 40.28                                  & 57.64                           & 0.00                               & \textbf{91.76}                & 35.62                            & 72.24                             & 0.09                                  & 84.63                             & 46.26                          & 76.76                               & 55.93                          & 64.17                            & 55.39                         & \textbf{44.28}                        & 70.40                        & 51.00                & 55.60                                \\
                                  & FSSAD + ours              & \textbf{94.92}               & 18.26                            & \textbf{59.87}                      & \textbf{91.63}                 & 62.24                                  & 62.92                           & 0.00                               & 89.32                         & \textbf{36.26}                   & \textbf{76.68}                    & \textbf{0.82}                         & \textbf{90.68}                    & \textbf{65.34}                 & \textbf{86.88}                      & \textbf{63.85}                 & \textbf{75.87}                   & \textbf{59.30}                & 39.08                                 & \textbf{93.14}               & \textbf{58.35}       & \textbf{56.59}                       \\ %
                                  & GFSS                      & 88.15                        & \textbf{11.19}                   & 17.43                               & 26.27                          & 23.81                                  & 35.30                           & \textbf{14.49}                     & 87.84                         & 28.81                            & 67.25                             & 0.00                                  & 82.40                             & 46.29                          & 74.47                               & 62.13                          & 55.09                            & 54.98                         & 31.11                                 & 53.82                        & 43.04                & 47.45                                \\
                                  & GFSS + ours               & \textbf{92.05}               & 3.69                             & \textbf{28.90}                      & \textbf{63.47}                 & \textbf{36.95}                         & \textbf{41.30}                  & 11.37                              & \textbf{90.14}                & \textbf{31.42}                   & \textbf{72.42}                    & \textbf{0.01}                         & \textbf{86.52}                    & \textbf{56.14}                 & \textbf{81.99}                      & \textbf{68.43}                 & \textbf{66.03}                   & \textbf{57.65}                & \textbf{39.09}                        & \textbf{79.21}               & \textbf{50.34}       & \textbf{51.70}                       \\ %
                                  & LwF                       & 90.75                        & 6.02                             & 2.30                                & 61.10                          & 41.41                                  & 56.14                           & 18.29                              & 84.85                         & 27.83                            & 67.91                             & 0.90                                  & 87.38                             & 54.79                          & \textbf{78.12}                      & 55.70                          & 67.77                            & 56.68                         & 41.21                                 & \textbf{65.75}               & 48.24                & 53.79                                \\
                                  & LwF + ours                & \textbf{94.51}               & \textbf{6.98}                    & \textbf{60.99}                      & \textbf{84.25}                 & \textbf{53.01}                         & \textbf{60.76}                  & \textbf{58.62}                     & \textbf{89.93}                & \textbf{35.03}                   & \textbf{73.86}                    & \textbf{1.13}                         & \textbf{89.12}                    & \textbf{60.44}                 & 68.25                               & \textbf{63.08}                 & \textbf{67.89}                   & \textbf{61.20}                & \textbf{42.41}                        & 0.02                         & \textbf{53.57}       & \textbf{62.17}                       \\ \hline
            \multirow{6}{*}{$5$}  & FSSAD                     & 92.50                        & \textbf{34.41}                   & 54.56                               & 59.23                          & 48.93                                  & 60.77                           & 0.00                               & \textbf{92.51}                & \textbf{39.62}                   & 73.16                             & \textbf{0.03}                         & 87.76                             & 55.17                          & 84.82                               & 62.15                          & \textbf{67.86}                   & 56.95                         & \textbf{41.07}                        & 89.71                        & 55.06                & 58.19                                \\
                                  & FSSAD + ours              & \textbf{94.89}               & 23.21                            & \textbf{60.65}                      & \textbf{91.12}                 & \textbf{60.76}                         & \textbf{61.97}                  & 0.00                               & 91.36                         & 38.48                            & \textbf{69.56}                    & 0.00                                  & \textbf{90.22}                    & \textbf{67.09}                 & \textbf{87.64}                      & \textbf{68.57}                 & 64.44                            & \textbf{59.19}                & 34.82                                 & \textbf{93.06}               & \textbf{57.85}       & \textbf{58.46}                       \\ %
                                  & GFSS                      & 91.95                        & \textbf{6.38}                    & 20.51                               & 41.12                          & \textbf{39.61}                         & 32.50                           & \textbf{15.34}                     & 89.13                         & \textbf{31.74}                   & 74.38                             & 0.00                                  & 86.33                             & 56.58                          & 83.95                               & 63.63                          & 62.60                            & 53.64                         & 36.00                                 & 82.77                        & 48.41                & 50.06                                \\
                                  & GFSS + ours               & \textbf{92.82}               & 0.68                             & \textbf{34.55}                      & \textbf{70.88}                 & 38.04                                  & \textbf{46.79}                  & 10.25                              & \textbf{89.85}                & 29.58                            & \textbf{74.83}                    & 0.00                                  & \textbf{91.55}                    & \textbf{68.62}                 & \textbf{88.06}                      & \textbf{68.27}                 & \textbf{73.45}                   & \textbf{57.04}                & \textbf{38.41}                        & \textbf{91.03}               & \textbf{53.24}       & \textbf{53.43}                       \\ %
                                  & LwF                       & 94.32                        & \textbf{15.46}                   & 58.28                               & 34.11                          & 53.06                                  & \textbf{63.98}                  & 5.56                               & \textbf{92.49}                & \textbf{34.83}                   & \textbf{75.77}                    & 0.19                                  & 88.38                             & 58.01                          & 85.28                               & 65.84                          & \textbf{68.98}                   & 58.14                         & \textbf{42.72}                        & 90.52                        & 54.30                & 56.44                                \\
                                  & LwF + ours                & \textbf{94.96}               & 9.65                             & \textbf{64.47}                      & \textbf{85.20}                 & \textbf{67.72}                         & 63.60                           & \textbf{20.80}                     & 92.04                         & 32.54                            & 74.83                             & \textbf{0.53}                         & \textbf{89.41}                    & \textbf{61.27}                 & \textbf{87.29}                      & \textbf{66.16}                 & 66.87                            & \textbf{59.38}                & 40.42                                 & \textbf{96.32}               & \textbf{58.67}       & \textbf{61.64}                       \\ \hline
            \multirow{6}{*}{$10$} & FSSAD                     & 92.83                        & \textbf{34.55}                   & 55.17                               & 70.26                          & 55.57                                  & 58.43                           & 28.24                              & \textbf{93.04}                & \textbf{36.19}                   & \textbf{76.88}                    & 0.00                                  & 89.69                             & 63.40                          & 87.49                               & 63.47                          & 74.04                            & 58.61                         & \textbf{38.28}                        & 93.57                        & 58.49                & 60.55                                \\
                                  & FSSAD + ours              & \textbf{94.69}               & 28.80                            & \textbf{71.23}                      & \textbf{86.53}                 & \textbf{57.96}                         & \textbf{61.27}                  & \textbf{32.34}                     & 90.53                         & 15.74                            & 74.71                             & \textbf{0.03}                         & \textbf{91.66}                    & \textbf{70.04}                 & \textbf{89.53}                      & \textbf{70.10}                 & \textbf{75.91}                   & \textbf{59.45}                & 30.38                                 & \textbf{95.71}               & \textbf{59.83}       & \textbf{61.85}                       \\ %
                                  & GFSS                      & 91.12                        & \textbf{5.28}                    & \textbf{27.23}                      & 46.22                          & 47.64                                  & 43.49                           & 30.08                              & 89.25                         & \textbf{32.52}                   & 74.37                             & 0.00                                  & 86.27                             & 56.06                          & 83.20                               & 63.35                          & 68.48                            & 54.79                         & \textbf{36.81}                        & 87.47                        & 51.18                & 51.25                                \\
                                  & GFSS + ours               & \textbf{94.30}               & 0.09                             & 25.81                               & \textbf{87.86}                 & \textbf{50.84}                         & \textbf{53.23}                  & \textbf{37.78}                     & \textbf{90.73}                & 21.54                            & \textbf{77.80}                    & 0.00                                  & \textbf{91.75}                    & \textbf{70.88}                 & \textbf{88.46}                      & \textbf{67.75}                 & \textbf{77.62}                   & \textbf{57.49}                & 35.11                                 & \textbf{92.22}               & \textbf{56.06}       & \textbf{55.11}                       \\ %
                                  & LwF                       & 94.15                        & \textbf{12.88}                   & 50.92                               & 71.65                          & 55.71                                  & 63.95                           & 46.94                              & \textbf{92.69}                & 34.84                            & \textbf{77.68}                    & 0.00                                  & 89.61                             & 61.17                          & 86.88                               & 67.11                          & \textbf{72.93}                   & 59.30                         & \textbf{39.91}                        & 94.63                        & 58.65                & 60.79                                \\
                                  & LwF + ours                & \textbf{96.49}               & 3.97                             & \textbf{65.69}                      & \textbf{87.96}                 & \textbf{72.94}                         & \textbf{69.08}                  & \textbf{54.31}                     & 90.81                         & \textbf{37.26}                   & 75.77                             & 0.00                                  & \textbf{92.17}                    & \textbf{68.33}                 & \textbf{87.46}                      & \textbf{67.54}                 & 70.44                            & \textbf{59.85}                & 36.60                                 & \textbf{95.99}               & \textbf{61.63}       & \textbf{63.47}                       \\ \hline
        \end{tabular}
    }
    \label{table:FSS_SemanticKITTI_fog}
\end{table*}

\begin{figure*}[t!]
    \centering
    \includegraphics[width=1\textwidth]{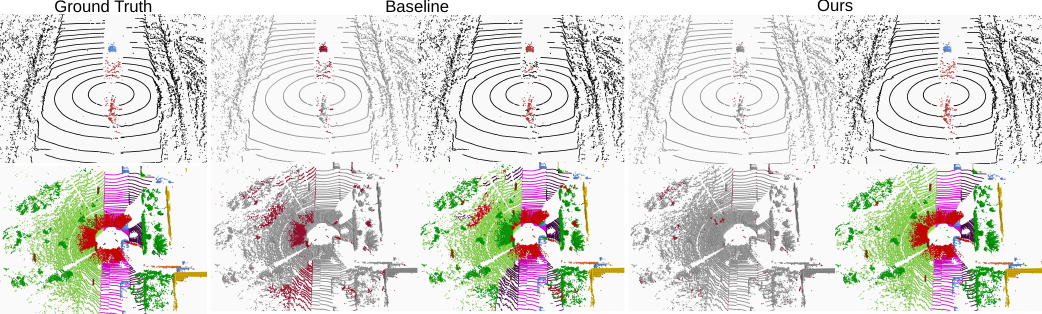}
    \caption{ Qualitative comparison results on different datasets.
    The results refer to FSSAD (Baseline) and FSSAD trained with our method (Ours).
    All results are obtained with $K=1$.
    For each method, we show on the left a binary-coded visualization of $\color{bin-correct}{\bullet}~$\textit{correctly} and $\color{bin-wrong}{\bullet}~$\textit{incorrectly} segmented points and on the right the class predictions. 
    The top row refers to the SemanticSpray dataset, with colormap:  
    $\color{SS-background}{\bullet}~$\textit{background}
$\color{SS-car}{\bullet}~$\textit{car}
$\color{SS-spray}{\bullet}~$\textit{spray}.
    The bottom row refers to the SemanticKITTI-fog dataset with colormap:
    $\color{car}{\bullet}~$\textit{car}
    $\color{road}{\bullet}~$\textit{road}
    $\color{sidewalk}{\bullet}~$\textit{sidewalk}
    $\color{other-ground}{\bullet}~$\textit{other-ground}
    $\color{building}{\bullet}~$\textit{building}
    $\color{fence}{\bullet}~$\textit{fence}
    $\color{vegetation}{\bullet}~$\textit{vegetation}
    $\color{trunk}{\bullet}~$\textit{trunk}
    $\color{terrain}{\bullet}~$\textit{terrain}
    $\color{fog}{\bullet}~$\textit{fog}.
    }
    \label{Fig:qualitative_res}
\end{figure*}

\textbf{FSS Performances.}
In Tables~\ref{table:FSS_SemanticSpray},~\ref{table:FSS_WADS},~\ref{table:FSS_SemanticKITTI_fog}, we report the comparison between the FSS baselines and the FSS baselines trained with our method.
We can observe that our approach greatly improves the segmentation performances of all baselines.
For example, the mIoU of FSSAD on the SemanticSpray dataset with $K=1$ is improved by $+28.19\%$ mIoU points.
The gain in performance can be attributed to an improvement in both base and novel class segmentation.
In fact, in addition to data scarcity, the task of FSS in adverse weather has the additional challenge of domain adaptation from source to target dataset. 
When measured in adverse weather conditions, base classes such as vehicles, pedestrians, or buildings can have drastically different properties (e.g., point density, intensity, measurement pattern) than when measured in good weather.
Our approach addresses both problems simultaneously by generating pseudo-labels for the entire LiDAR scan, including instances of both base and novel classes.
When we compare our approach against the fully supervised trained methods, we see that we can achieve competitive performances.
For example, on the SemanticSpray dataset with FSSAD and $K=10$, our method has a difference of only $1.15\%$ mIoU points compared to the fully supervised approach. 
This is achieved while using only $10$ labeled scans, whereas the fully supervised method uses more than  $\si{8}{k}$ scans.
In some cases, such as the WADS dataset, we see that our approach even outperforms the fully supervised method.
This result is mainly due to a better performance on the base classes, which is also contributed by training on a large source dataset in Stage Zero. 
In the specific case of the WADS dataset, the data used for the fully supervised setting is relatively small and may not be sufficient to achieve high performance in all classes.
In the SemanticKITTI-fog dataset, although our method improves on the baseline, the resulting performance still lags behind the fully supervised method. 
In Fig.~\ref{Fig:qualitative_res}, we show a qualitative comparison of the baseline and our method.

\textbf{Performance on Good Weather Data.}
From Table~\ref{table:FSS_SemanticSpray},~\ref{table:FSS_WADS} and~\ref{table:FSS_SemanticKITTI_fog}, we can also observe that our method allows for a much higher retention of performance on the source dataset than the baselines.
For example, GFSS trained on the WADS dataset with $K=10$ improves the performance by $+33.58\%$ mIoU points. 
We highlight that compared to our method, the baselines do not use any data from $\mathcal{D}_s$ during FSS training but aim to retain performance in other ways (e.g., knowledge distillation~\cite{10160674, li2017learning}).
As we can see from the result, this is often not enough, especially when the target dataset only contains a subset of the base classes (e.g., SemanticSpray).
However, since the ground truth data from $\mathcal{D}_s$ is a prerequisite for training the FSS base models, using it in training does not add any additional labeling effort to the task but greatly improves performance.

\textbf{Comparison with SSL Baselines.}
\begin{table}[t!]
    \centering
    \caption{Comparison with other SSL methods.
    Results on the SemanticSpray dataset with $K=5$ using as baseline LwF~\cite{li2017learning}. 
    The results of our method are reported when using Stage One of our method (Stage Zero is also used to derive $\phi_{S_0})$ in combination with LwF.
    The other SSL baselines are jointly trained with the LwF objective function.
    }
    \resizebox{0.9\columnwidth}{!}{%

        \begin{tabular}{@{}llccc@{}}
            \toprule
            Dataset                             & Method              & mIoU~$C_b$     & mIoU~$C_n$     & mIoU           \\ \midrule
            \multirow{4}{*}{SemanticSpray}      & \textit{LwF Only}   & 83.06          & 73.85          & 79.99          \\
                                                & GuidedPointContrast & 85.18          & 72.10          & 80.82          \\
                                                & PointContrast       & 84.51          & 72.81          & 80.61          \\
                                                & Ours               & \textbf{97.34} & \textbf{82.75} & \textbf{92.48} \\ \midrule
            \multirow{4}{*}{WADS}               & \textit{LwF Only}   & 21.02          & 76.25          & 24.47          \\
                                                & GuidedPointContrast & 20.28          & 62.91          & 22.95          \\
                                                & PointContrast       & 21.07          & 75.07          & 24.45          \\
                                                & Ours                 & \textbf{22.22} & \textbf{79.78} & \textbf{25.82} \\ \midrule
            \multirow{4}{*}{SemanticKITTI-fog} & \textit{LwF Only}   & 52.39          & 90.52          & 54.30          \\
                                                & GuidedPointContrast & 52.95          & 91.80          & 54.89          \\
                                                & PointContrast       & 52.07          & 91.17          & 54.03          \\
                                                & Ours                & \textbf{54.80} & \textbf{94.53} & \textbf{56.78} \\ \bottomrule
        \end{tabular}

    }
    \label{table:SSL_eval}
\end{table}

In Table~\ref{table:SSL_eval}, we report the comparison results between our proposed method and other state-of-the-art approaches for SSL in outdoor 3D point clouds.
From the results we can see that our approach performs better than the other SSL methods. 
For example, in the SemanticSpray dataset, we improve performance against the second-best method (GuidedPointContrast) by $+11.66\%$ mIoU points.

\textbf{Limitations.}
Although our method can improve the performance of the current state-of-the-art FSS methods, it still shares some of the limitations of FSS and SSL learning.
For example, class-wise performance depends on the semantic classes contained in the $K$ labeled data using training.
If the $K$ labeled scans do not contain any instances of a specific class, the resulting performance will be low.
For instance, this can be observed on the \textit{other-ground} class performance for the SemanticKITTI-fog dataset, where all methods perform much lower than in the fully-supervised approach.
An additional limitation is the risk of training with wrong pseudo-labels.
By integrating incorrect labels during training, the performance of specific classes could be reduced. 
This can be observed for the class \textit{bicycle} of the SemanticKITTI-fog dataset, where the performance of the baseline methods is higher than our approach.
Therefore, a careful evaluation of the models trained with pseudo-labels should be performed in order to check for undesired learned biases.

\subsection{Ablation Studies}
\begin{table}[t!]
    \centering
    \caption{Ablation study on each component contribution evaluated on SemanticSpray using LwF~\cite{li2017learning} and $K=5$.
        mIoU~$C_b$ and mIoU~$C_n$ refer to the base and novel classes respectively.
        mIoU~$\mathcal{D}_s$ refers to the source dataset (nuScenes).
        BMS refers to the best model selection.
        Mix to the data mixing operation.
        SCE, when true means that the symmetric cross entropy loss was used, when false that the cross entropy is applied.
        $\omega_0$ refers to the use of the warmup parameter described in \Cref{sec:stage_one}.
    }
    \resizebox{1\columnwidth}{!}{%
        \begin{tabular}{@{}cccccccccc@{}}
            \toprule
            $S_1$  & $S_2$  & BMS    & Mix    & SCE    & $\omega_0$ & mIoU~$C_b$ & mIoU~$C_n$ & mIoU  & mIoU~$\mathcal{D}_s$ \\ \midrule
            \cmark & -      & \xmark & -      & \cmark & \cmark     & 86.99      & 63.96      & 79.32 & 9.42                 \\
            \cmark & -      & \cmark & -      & \xmark & \cmark     & 71.72      & 77.7       & 73.71 & 2.6                  \\
            \cmark & -      & \cmark & -      & \cmark & \xmark     & 93.23      & 64.61      & 83.69 & 2.74                 \\
            \cmark & -      & \cmark & -      & \cmark & \cmark     & 97.34      & 82.75      & 92.48 & 7.29                 \\
            \cmark & \cmark & \xmark & \cmark & \cmark & \cmark     & 87.23      & 64.11      & 79.53 & 9.53                 \\
            \cmark & \cmark & \cmark & \xmark & \cmark & \cmark     & 97.20      & 82.73      & 92.38 & 6.99                 \\
            \cmark & \cmark & \cmark & \cmark & \cmark & \cmark     & 96.91      & 80.93      & 91.58 & 62.89                \\ \bottomrule
        \end{tabular}

    }
    \label{table:Ablation_component_eval}
\end{table}

\textbf{Component Contribution.}
In Table~\ref{table:Ablation_component_eval}, we report the contribution of each component of our approach.
We can see that while training with only Stage One yields better results on the target dataset, the performance retention on the source dataset is much lower.
The selection of the best model on the pseudo-validation is fundamental to the first stage, where it helps to improve the quality of the generated pseudo-labels and lowers the risk of overfitting.
In Stage Two, we see that the best model selection and the data mix operation are necessary to achieve high performance on both source and target datasets.

\textbf{Wrong Pseudo-Label Influence.}
In Table~\ref{table:Ablation_component_eval}, we can also observe that the best model selection combined with the symmetric cross entropy and the warmup parameter $\omega_0$ contribute to the overall method robustness to wrong pseudo labels. 
In particular, best model selection allows to reject models for pseudo-label generation that are sub-optimal with respect to the best-trained model, thus maintaining or improving the quality of the pseudo-labels.
The symmetric cross entropy also improves the performance since it is designed to be more robust to the cross entropy loss to noisy labels.
Finally, the warmup parameter $\omega_0$ is useful to avoid the use of early-stage noisy pseudo-labels, which can lead to an overall performance degradation in the entire optimization process.

\textbf{Sensitivity to Ground Truth Labels.}
\begin{table}[t!]
    \centering
    \caption{Ablation study on the sensitivity of the method to the number of $C_n$ points in the target dataset.
    For each value of $K$, $5$ runs are performed using an average of \SIrange[range-phrase=--, range-units=single]{1}{10}{\kilo{}} $C_n$ points.
    The results refer to both stages of our method applied to LwF on SemanticKITTI-fog.
    }
    \resizebox{0.8\columnwidth}{!}{%
\begin{tabular}{cccccc}
\toprule
K                   &    ~   & mIoU~$C_b$ & mIoU~$C_n$ & mIoU  & mIoU~$\mathcal{D}_s$ \\ \midrule
\multirow{2}{*}{1}  & \textit{mean}  & 51.85     & 63.71     & 52.44 & 56.80     \\
                    & \textit{stdev} & 4.02      & 39.61     & 3.31  & 4.73      \\ \midrule
\multirow{2}{*}{5}  & \textit{mean}  & 56.16     & 92.46     & 57.98 & 60.26     \\
                    & \textit{stdev} & 2.73      & 9.17      & 2.66  & 3.55      \\ \midrule
\multirow{2}{*}{10} & \textit{mean}  & 59.07     & 97.04     & 60.97 & 62.41     \\
                    & \textit{stdev} & 1.42      & 0.44      & 1.35  & 1.54      \\ \bottomrule
\end{tabular}
}
\label{tab:sensitivity_to_Cn}
\end{table}

In Tab.~\ref{tab:sensitivity_to_Cn}, we report the sensitivity of the proposed method to the number of novel points $C_n$ present in the ground truth labeled dataset $\mathcal{D}_t^\ell$.
We perform $5$ experiments for each value of $K$, choosing the scans of   $\mathcal{D}_t^\ell$ so that we have a different average of novel points (from \SIrange[range-phrase=--, range-units=single]{1}{10}{\kilo{}}, with $\SI{2.5}{\kilo{}}$ increments), and then compute the mean and standard deviation (stdev) among the different mIoU values.
As expected, the mean values are higher for larger values of $K$.
However, we can also notice that stdev of values for $K=1$ is much higher than for larger values $K$, indicating that the method is much more sensitive to the content of  $\mathcal{D}_t^\ell$ when training with fewer labels.

\section{Conclusion} 
\label{sec:conclusion}
In this paper, we present a simple yet effective approach for semantic segmentation of LiDAR scans in adverse weather conditions.
Our method allows for a label-efficient training pipeline that relies on only a few annotated scans in adverse weather. 
We combine few-shot semantic segmentation (FSS) and semi-supervised learning (SSL) to expand the available pool of training data by generating pseudo-labels for unlabelled data.
Additionally, we incorporate good weather data into our training pipeline so that our models perform well in both good and bad weather conditions.
Results on several datasets show that our method is highly effective in segmenting adverse weather effects such as spray, snow, and fog, achieving competitive performance with fully supervised methods while using only a fraction of the data.

\bibliographystyle{IEEEtran}
\bibliography{mybib}

\begin{thebibliography}{10}
\providecommand{\url}[1]{#1}
\csname url@rmstyle\endcsname
\providecommand{\newblock}{\relax}
\providecommand{\bibinfo}[2]{#2}
\providecommand\BIBentrySTDinterwordspacing{\spaceskip=0pt\relax}
\providecommand\BIBentryALTinterwordstretchfactor{4}
\providecommand\BIBentryALTinterwordspacing{\spaceskip=\fontdimen2\font plus
\BIBentryALTinterwordstretchfactor\fontdimen3\font minus \fontdimen4\font\relax}
\providecommand\BIBforeignlanguage[2]{{%
\expandafter\ifx\csname l@#1\endcsname\relax
\typeout{** WARNING: IEEEtran.bst: No hyphenation pattern has been}%
\typeout{** loaded for the language `#1'. Using the pattern for}%
\typeout{** the default language instead.}%
\else
\language=\csname l@#1\endcsname
\fi
#2}}

\bibitem{heinzler2020cnn}
R.~Heinzler, F.~Piewak, P.~Schindler, and W.~Stork, ``Cnn-based lidar point cloud de-noising in adverse weather,'' \emph{IEEE Robotics and Automation Letters}, vol.~5, no.~2, pp. 2514--2521, 2020.

\bibitem{seppanen20224denoisenet}
A.~Sepp{\"a}nen, R.~Ojala, and K.~Tammi, ``4denoisenet: Adverse weather denoising from adjacent point clouds,'' \emph{IEEE Robotics and Automation Letters}, vol.~8, no.~1, pp. 456--463, 2022.

\bibitem{piroli2023energy}
A.~Piroli, V.~Dallabetta, J.~Kopp, M.~Walessa, D.~Meissner, and K.~Dietmayer, ``Energy-based detection of adverse weather effects in lidar data,'' \emph{IEEE Robotics and Automation Letters}, 2023.

\bibitem{xiao20233d}
A.~Xiao, J.~Huang, W.~Xuan, R.~Ren, K.~Liu, D.~Guan, A.~E. Saddik, S.~Lu, and E.~Xing, ``3d semantic segmentation in the wild: Learning generalized models for adverse-condition point clouds,'' \emph{arXiv preprint arXiv:2304.00690}, 2023.

\bibitem{kurup2021dsor}
A.~Kurup and J.~Bos, ``Dsor: A scalable statistical filter for removing falling snow from lidar point clouds in severe winter weather,'' \emph{arXiv preprint arXiv:2109.07078}, 2021.

\bibitem{kong2023robo3d}
L.~Kong, Y.~Liu, X.~Li, R.~Chen, W.~Zhang, J.~Ren, L.~Pan, K.~Chen, and Z.~Liu, ``Robo3d: Towards robust and reliable 3d perception against corruptions,'' in \emph{Proceedings of the IEEE/CVF International Conference on Computer Vision}, 2023, pp. 19\,994--20\,006.

\bibitem{qi2017pointnet++}
C.~R. Qi, L.~Yi, H.~Su, and L.~J. Guibas, ``Pointnet++: Deep hierarchical feature learning on point sets in a metric space,'' \emph{Advances in neural information processing systems}, vol.~30, 2017.

\bibitem{choy20194d}
C.~Choy, J.~Gwak, and S.~Savarese, ``4d spatio-temporal convnets: Minkowski convolutional neural networks,'' in \emph{Proceedings of the IEEE/CVF conference on computer vision and pattern recognition}, 2019, pp. 3075--3084.

\bibitem{tang2020searching}
H.~Tang, Z.~Liu, S.~Zhao, Y.~Lin, J.~Lin, H.~Wang, and S.~Han, ``Searching efficient 3d architectures with sparse point-voxel convolution,'' in \emph{European conference on computer vision}.\hskip 1em plus 0.5em minus 0.4em\relax Springer, 2020, pp. 685--702.

\bibitem{dreissig2023survey}
M.~Dreissig, D.~Scheuble, F.~Piewak, and J.~Boedecker, ``Survey on lidar perception in adverse weather conditions,'' \emph{arXiv preprint arXiv:2304.06312}, 2023.

\bibitem{piroli2022detection}
A.~Piroli, V.~Dallabetta, M.~Walessa, D.~Meissner, J.~Kopp, and K.~Dietmayer, ``Detection of condensed vehicle gas exhaust in lidar point clouds,'' in \emph{2022 IEEE 25th International Conference on Intelligent Transportation Systems (ITSC)}.\hskip 1em plus 0.5em minus 0.4em\relax IEEE, 2022, pp. 600--606.

\bibitem{piroli2023towards}
A.~Piroli, V.~Dallabetta, J.~Kopp, M.~Walessa, D.~Meissner, and K.~Dietmayer, ``Towards robust 3d object detection in rainy conditions,'' in \emph{2023 IEEE 26th International Conference on Intelligent Transportation Systems (ITSC)}.\hskip 1em plus 0.5em minus 0.4em\relax IEEE, 2023, pp. 3471--3477.

\bibitem{catalano2023few}
N.~Catalano and M.~Matteucci, ``Few shot semantic segmentation: a review of methodologies and open challenges,'' \emph{arXiv preprint arXiv:2304.05832}, 2023.

\bibitem{10160674}
J.~Mei, J.~Zhou, and Y.~Hu, ``Few-shot 3d lidar semantic segmentation for autonomous driving,'' in \emph{2023 IEEE International Conference on Robotics and Automation (ICRA)}, 2023, pp. 9324--9330.

\bibitem{li2017learning}
Z.~Li and D.~Hoiem, ``Learning without forgetting,'' \emph{IEEE transactions on pattern analysis and machine intelligence}, vol.~40, no.~12, pp. 2935--2947, 2017.

\bibitem{myers2021generalized}
J.~Myers-Dean, Y.~Zhao, B.~Price, S.~Cohen, and D.~Gurari, ``Generalized few-shot semantic segmentation: All you need is fine-tuning,'' \emph{arXiv preprint arXiv:2112.10982}, 2021.

\bibitem{liu2022less}
M.~Liu, Y.~Zhou, C.~R. Qi, B.~Gong, H.~Su, and D.~Anguelov, ``Less: Label-efficient semantic segmentation for lidar point clouds,'' in \emph{European Conference on Computer Vision}.\hskip 1em plus 0.5em minus 0.4em\relax Springer, 2022, pp. 70--89.

\bibitem{yang2022survey}
X.~Yang, Z.~Song, I.~King, and Z.~Xu, ``A survey on deep semi-supervised learning,'' \emph{IEEE Transactions on Knowledge and Data Engineering}, 2022.

\bibitem{li2019learning}
X.~Li, Q.~Sun, Y.~Liu, Q.~Zhou, S.~Zheng, T.-S. Chua, and B.~Schiele, ``Learning to self-train for semi-supervised few-shot classification,'' \emph{Advances in neural information processing systems}, vol.~32, 2019.

\bibitem{wang2019symmetric}
Y.~Wang, X.~Ma, Z.~Chen, Y.~Luo, J.~Yi, and J.~Bailey, ``Symmetric cross entropy for robust learning with noisy labels,'' in \emph{Proceedings of the IEEE/CVF international conference on computer vision}, 2019, pp. 322--330.

\bibitem{xie2020pointcontrast}
S.~Xie, J.~Gu, D.~Guo, C.~R. Qi, L.~Guibas, and O.~Litany, ``Pointcontrast: Unsupervised pre-training for 3d point cloud understanding,'' in \emph{Computer Vision--ECCV 2020: 16th European Conference, Glasgow, UK, August 23--28, 2020, Proceedings, Part III 16}.\hskip 1em plus 0.5em minus 0.4em\relax Springer, 2020, pp. 574--591.

\bibitem{oord2018representation}
A.~v.~d. Oord, Y.~Li, and O.~Vinyals, ``Representation learning with contrastive predictive coding,'' \emph{arXiv preprint arXiv:1807.03748}, 2018.

\bibitem{jiang2021guided}
L.~Jiang, S.~Shi, Z.~Tian, X.~Lai, S.~Liu, C.-W. Fu, and J.~Jia, ``Guided point contrastive learning for semi-supervised point cloud semantic segmentation,'' in \emph{Proceedings of the IEEE/CVF international conference on computer vision}, 2021, pp. 6423--6432.

\bibitem{hinton2015distilling}
G.~Hinton, O.~Vinyals, and J.~Dean, ``Distilling the knowledge in a neural network,'' \emph{arXiv preprint arXiv:1503.02531}, 2015.

\bibitem{xiao2022polarmix}
A.~Xiao, J.~Huang, D.~Guan, K.~Cui, S.~Lu, and L.~Shao, ``Polarmix: A general data augmentation technique for lidar point clouds,'' \emph{Advances in Neural Information Processing Systems}, vol.~35, pp. 11\,035--11\,048, 2022.

\bibitem{behley2019semantickitti}
J.~Behley, M.~Garbade, A.~Milioto, J.~Quenzel, S.~Behnke, C.~Stachniss, and J.~Gall, ``Semantickitti: A dataset for semantic scene understanding of lidar sequences,'' in \emph{Proceedings of the IEEE/CVF international conference on computer vision}, 2019, pp. 9297--9307.

\bibitem{hahner2021fog}
M.~Hahner, C.~Sakaridis, D.~Dai, and L.~Van~Gool, ``Fog simulation on real lidar point clouds for 3d object detection in adverse weather,'' in \emph{Proceedings of the IEEE/CVF International Conference on Computer Vision}, 2021, pp. 15\,283--15\,292.

\bibitem{caesar2020nuscenes}
H.~Caesar, V.~Bankiti, A.~H. Lang, S.~Vora, V.~E. Liong, Q.~Xu, A.~Krishnan, Y.~Pan, G.~Baldan, and O.~Beijbom, ``nuscenes: A multimodal dataset for autonomous driving,'' in \emph{Proceedings of the IEEE/CVF conference on computer vision and pattern recognition}, 2020, pp. 11\,621--11\,631.

\end{thebibliography}

\end{document}